\documentclass[runningheads]{llncs}

\usepackage[FINAL,year=2026,ID=*****]{eccv}

\usepackage{eccvabbrv}

\usepackage{graphicx}
\usepackage{booktabs}

\usepackage[accsupp]{axessibility}  

\usepackage{hyperref}

\usepackage{orcidlink}

\begin{document}

\title{MLLM-Assisted Audio VOS: A 3rd Place Report for the MeViS-Audio Track, 8th LSVOS Challenge} 

\titlerunning{ 3rd Place Report for the MeViS-Audio Track, 8th LSVOS Challenge}

\author{Liangtao Shi\inst{1} \and
Jinxia Xie\inst{2} \and
Xiantao Hu\inst{2}  \and
Ting Liu\inst{3}}

\authorrunning{L. Shi et al.}

\institute{Hefei University of Technology
\and
Nanjing University of Science and Technology
\and
Hunan Police College}

\maketitle

\begin{abstract}
In this technical report, we present a training-free framework for audio-guided video object segmentation, which integrates Multimodal Large Language Models (MLLMs) with SAM-based segmentation models.
We decompose the task into several stages and identify suitable foundation models for each stage.
Without introducing additional model training or task-specific fine-tuning, our approach leverages the strong multimodal reasoning capabilities of MLLMs to model text-visual correspondence and employs SAM-based models for accurate object mask generation. 
The proposed framework demonstrates the effectiveness of leveraging foundation models for audio-guided video segmentation and achieves competitive performance in the MeViS-Audio Track of the 8th LSVOS Challenge.
\end{abstract}

\section{Introduction}
\label{sec:intro}
The 8th Large-scale Video Object Segmentation (LSVOS) Challenge provides a series of challenging benchmarks for advancing video object segmentation. It evaluates the capability of modern segmentation systems to understand complex visual content, accurately localize target objects, and maintain robust pixel-level segmentation across diverse video scenarios.
With the rapid development of video understanding and multimodal learning, the challenge further explores diverse segmentation settings that incorporate visual, language, and audio cues through three tracks, covering complex video object segmentation, text-based referring video object segmentation, and audio-guided video object segmentation.

1) \textbf{Complex Video Object Segmentation (MOSEv2 Track):} 
As a classic track of the LSVOS Challenge, this track focuses on robust video object segmentation under challenging scenarios, including object disappearance and reappearance, severe occlusion, small objects, and crowded scenes.
The evaluation is conducted on a test subset of MOSEv2~\cite{ding2025mosev2}, an extension of MOSE~\cite{ding2023mose} designed for complex VOS. 
MOSEv2 contains 5,024 videos with 10,074 objects and 701,976 masks across 200 categories, introducing more realistic challenges such as long-term disappearance, camouflage, adverse conditions, and non-physical targets. 

2) \textbf{Text-based Referring Video Object Segmentation (MeViS-Text Track):} 
This track aims to segment video objects specified by natural-language motion expressions. The challenge evaluates models on a subset of MeViSv2\cite{ding2025mevis}. 
MeViSv2 extends the original MeViS\cite{ding2023mevis} benchmark, which contains 2,006 videos, 8,171 objects, 28,570 motion expressions, and approximately 443K object masks. 
The new version preserves the original video collection while expanding the language annotations to 33,458 sentences and adding motion-reasoning and no-target expressions, as well as audio descriptions and trajectory annotations. 
The task requires jointly modeling language, object motion, and temporal context, making it a challenging testbed for motion-aware vision-language understanding and referring video object segmentation.

3) \textbf{Audio-guided Video Object Segmentation (MeViS-Audio Track):} 
The 8th LSVOS Challenge introduces this new track to extend referring expression video segmentation from language to audio. 
Given an audio clip associated with the target object, models are required to localize and segment the corresponding object throughout the video. 
The task involves associating acoustic cues with visual objects and maintaining the correspondence over time, introducing an additional challenge of audio-visual alignment beyond conventional referring video object segmentation. 
The evaluation is performed on a subset of MeViS-Audio\cite{ding2025mevis}.

Although the MeViS-Audio Track is newly introduced, previous solutions from the MOSEv2 and MeViS-Text Tracks~\cite{liu2025lsvos} provide valuable insights. 
Meanwhile, recent studies on Audio-guided Video Object Segmentation, including the 5th PVUW Challenge~\cite{liu2026report}, have explored effective solutions. 
A notable paradigm is to convert audio into text via ASR, allowing existing RVOS models and MLLMs to be applied. 
Following this paradigm, our framework first converts audio into textual descriptions using ASR models. 
The MLLM then jointly reasons over textual and visual inputs to refine referring prompts and select informative keyframes. 
RVOS/VOS models subsequently perform pixel-level mask prediction, followed by MLLM-based semantic verification between the predicted masks and textual descriptions. 
This framework achieves 3rd place in the MeViS-Audio Track of the 8th LSVOS Challenge.

\section{Method}
Our method develops a training-free pipeline for Audio-guided Video Object Segmentation, where MLLMs perform video-text understanding, while SAM-based models perform pixel-level segmentation.
The overall pipeline comprises four stages: (1) audio-to-text conversion, (2) video-text joint analysis, (3) text-based video segmentation and mask-based tracking, and (4) mask-text consistency verification.

\subsection{Stage 1: Audio-to-Text Conversion.}
Given an audio clip, we first convert it into a textual representation using an automatic speech recognition (ASR) model. 
Specifically, we adopt \textit{Qwen3-ASR-1.7B} ~\cite{shi2026qwen3} to transcribe the input audio \(A\):
\[
q = \Phi_{\mathrm{ASR}}(A),
\]
where \(q\) denotes the generated transcription.
This modality conversion enables the audio-guided task to leverage existing text-conditioned video understanding and segmentation models without requiring additional task-specific training. 
However, the ASR transcription alone may provide only a coarse-grained description of the referred targets, which may be insufficient for existing RVOS methods when multiple targets, fine-grained attributes, or complex spatial and temporal cues are involved. 
Therefore, we employ an MLLM to perform joint analysis of the textual query and video content, refining the original query into a more informative and fine-grained referring prompt for subsequent video object segmentation.

\subsection{Stage 2: Video-Text Joint Analysis.}
Given the transcribed query q and the video V, we employ \textit{Gemini-3-Flash-Preview} to jointly analyze the textual and visual information and identify the referred instances. 
Specifically, we determine the number of target instances and generate an instance-specific referring prompt and a representative keyframe for each target.
For the $i$-th target, we obtain a structured tuple
\[
o_i = (d_i, k_i),
\]
where $d_i$ denotes a fine-grained referring prompt and $k_i$ denotes the selected keyframe index. 
The prompt captures the visual characteristics and contextual cues that distinguish the target from other instances, while the keyframe indexed by $k_i$ serves as the initialization frame for subsequent tracking.

\subsection{Stage 3: Text-based Video Segmentation and Mask-based Object Tracking.}
Given the target description \(d_i\), we first apply MomentSeg~\cite{dai2025momentseg} to perform referring video object segmentation on the entire video. For each target instance \(o_i\), MomentSeg predicts a coarse mask sequence:
\[
\tilde{M}_i = \{\tilde{m}_{i,t}\}_{t=1}^{T}
= \Phi_{\mathrm{MomentSeg}}(V, d_i),
\]
where $\tilde{m}_{i,t}$ denotes the predicted mask at frame $t$. 
While MomentSeg provides coarse masks over the entire video, these masks may lack the pixel-level accuracy and temporal consistency required for high-quality segmentation. 
We therefore use the predicted mask at the MLLM-selected keyframe $k_i$, denoted as $\tilde{m}_{i,k_i}$, as the initial mask for subsequent bidirectional propagation and refinement with DAM4SAM~\cite{videnovic2025distractor}.
Starting from the keyframe $k_i$, DAM4SAM propagates and refines the target mask bidirectionally across the video:
\[
\hat{M}_i = \{\hat{m}_{i,t}\}_{t=1}^{T}
= \Phi_{\mathrm{DAM4SAM}}(V, \tilde{m}_{i,k_i}, k_i).
\]
For multiple target instances, the above procedure is performed independently for each instance, using its corresponding referring description and keyframe mask.
The resulting per-instance mask sequences are subsequently aggregated to obtain the final segmentation result.

\subsection{Stage 4: Mask-Text Consistency Verification.}
Although DAM4SAM provides strong pixel-level segmentation and temporal tracking capabilities, its propagated masks may still drift toward visually similar distractors or fail under occlusion. 
To mitigate these errors, we introduce \textit{Gemini-3-Flash-Preview} as a semantic verification module. Specifically, we merge the predicted masks of all instances in each frame and overlay the resulting mask on the corresponding video frame. 
The visualized masks, together with the transcribed text q, are then fed to \textit{Gemini-3-Flash-Preview} to verify their semantic consistency. 
If the predicted mask is inconsistent with the query, it is discarded by setting the mask to empty; otherwise, it is retained.

\section{Experiments}
\subsection{Evaluation Metrics}
Following the official evaluation protocol of the 8th LSVOS MeViS Audio challenge, we adopt five metrics for evaluation, including region similarity ($\mathcal{J}$), boundary accuracy ($\mathcal{F}$), N-acc., T-acc., and the final score. The region similarity metric $\mathcal{J}$ measures the overlap between the predicted segmentation regions and the ground-truth regions. The boundary accuracy metric $\mathcal{F}$ evaluates the alignment between the predicted object boundaries and the ground-truth boundaries. The combined metric $\mathcal{J}\&\mathcal{F}$ is computed as the average of $\mathcal{J}$ and $\mathcal{F}$. N-acc. and T-acc. evaluate the model performance on no-target and target cases, respectively. The final score is computed by combining $\mathcal{J}\&\mathcal{F}$, N-acc., and T-acc. for comprehensive evaluation.

\subsection{Ablation Study}

\begin{table}[t]
\centering
\caption{Comparison of different models on the 8th LSVOS MeViS-Audio test set.}
\label{tab:model_comparison}
\begin{tabular}{lcccccc}
\hline
Method & N-acc. & T-acc. & $\mathcal{J}$ & $\mathcal{F}$ & $\mathcal{J}\&\mathcal{F}$ & Final Score \\
\hline
MomentSeg & 41.38 & 92.53 & 49.72 & 56.47 & 53.09 & 62.33 \\
+ DAM4SAM & 41.38 & 92.53 & 53.81 & 59.63 & 56.72 & 63.54 \\
+ Consistency Verification & 96.55 & 57.83 & 44.34 & 47.71 & 46.03 & 66.80 \\
\hline
\end{tabular}
\end{table}

We evaluate the effectiveness of different components in our pipeline on the test set of the 8th LSVOS MeViS-Audio challenge. As shown in Table~\ref{tab:model_comparison}, MomentSeg achieves a $\mathcal{J}\&\mathcal{F}$ score of 53.09\%, providing reliable initial segmentation results based on text-guided prompts. By incorporating DAM4SAM for mask propagation and refinement, the $\mathcal{J}$, $\mathcal{F}$, and $\mathcal{J}\&\mathcal{F}$ scores improve to 53.81\%, 59.63\%, and 56.72\%, respectively, demonstrating the effectiveness of temporal mask refinement.

After introducing the consistency verification module, the final score further increases from 63.54\% to 66.80\%. Although the pixel-level segmentation metrics decrease due to stricter semantic filtering, N-acc. is significantly improved from 41.38\% to 96.55\%, indicating that the verification module effectively enhances the ability to distinguish target and no-target expressions. Overall, the results demonstrate that each component contributes differently to the final system, where DAM4SAM improves mask quality and consistency verification enhances the robustness of audio-guided video object segmentation.

\section{Conclusion}

In this work, we present a training-free framework for audio-guided video object segmentation by integrating multimodal large language models with SAM-based segmentation models. By combining audio understanding, multimodal reasoning, mask propagation, and semantic verification, our framework effectively aligns audio cues with visual targets and achieves temporally consistent segmentation without task-specific training. Experiments on the MeViS-Audio Track of the 8th LSVOS Challenge demonstrate the effectiveness of our approach, achieving competitive performance and ranking 3rd in the challenge. 

\bibliographystyle{splncs04}
\bibliography{main}

@String(CVPR  = {IEEE Conf. Comput. Vis. Pattern Recog.})

@String(ICCV  = {Int. Conf. Comput. Vis.})

@String(ECCV  = {Eur. Conf. Comput. Vis.})

@String(CVPR  = {CVPR})

@String(ICCV  = {ICCV})

@String(ECCV  = {ECCV})

@article{shi2026qwen3,
  title={Qwen3-asr technical report},
  author={Shi, Xian and Wang, Xiong and Guo, Zhifang and Wang, Yongqi and Zhang, Pei and Zhang, Xinyu and Guo, Zishan and Hao, Hongkun and Xi, Yu and Yang, Baosong and others},
  journal={arXiv preprint arXiv:2601.21337},
  year={2026}
}

@article{dai2025momentseg,
  title={MomentSeg: Moment-Centric Sampling for Enhanced Video Pixel Understanding},
  author={Dai, Ming and Yang, Sen and Duan, Boqiang and Yang, Wankou and Wang, Jingdong},
  journal={ECCV},
  year={2026}
}

@inproceedings{videnovic2025distractor,
  title={A distractor-aware memory for visual object tracking with sam2},
  author={Videnovic, Jovana and Lukezic, Alan and Kristan, Matej},
  booktitle={CVPR},
  pages={24255--24264},
  year={2025},
}

@inproceedings{ding2023mose,
  title={MOSE: A new dataset for video object segmentation in complex scenes},
  author={Ding, Henghui and Liu, Chang and He, Shuting and Jiang, Xudong and Torr, Philip HS and Bai, Song},
  booktitle={ICCV},
  pages={20167--20177},
  year={2023},
}

@article{ding2025mosev2,
  title={MOSEv2: A more challenging dataset for video object segmentation in complex scenes},
  author={Ding, Henghui and Ying, Kaining and Liu, Chang and He, Shuting and Jiang, Xudong and Jiang, Yu-Gang and Torr, Philip HS and Bai, Song},
  journal={arXiv preprint arXiv:2508.05630},
  year={2025}
}

@inproceedings{ding2023mevis,
  title={Mevis: A large-scale benchmark for video segmentation with motion expressions},
  author={Ding, Henghui and Liu, Chang and He, Shuting and Jiang, Xudong and Loy, Chen Change},
  booktitle={ICCV},
  pages={2694--2703},
  year={2023},
}

@article{ding2025mevis,
  title={MeViS: A multi-modal dataset for referring motion expression video segmentation},
  author={Ding, Henghui and Liu, Chang and He, Shuting and Ying, Kaining and Jiang, Xudong and Loy, Chen Change and Jiang, Yu-Gang},
  journal={IEEE TPAMI},
  year={2025},
  publisher={IEEE}
}

@article{liu2025lsvos,
  title={Lsvos 2025 challenge report: Recent advances in complex video object segmentation},
  author={Liu, Chang and Ding, Henghui and Ying, Kaining and Hong, Lingyi and Xu, Ning and Yang, Linjie and Fan, Yuchen and Gao, Mingqi and Chen, Jingkun and Miao, Yunqi and others},
  journal={arXiv preprint arXiv:2510.11063},
  year={2025}
}

@article{liu2026report,
  title={Report of the 5th PVUW Challenge: Towards More Diverse Modalities in Pixel-Level Understanding},
  author={Liu, Chang and Ding, Henghui and Ravi, Nikhila and Wei, Yunchao and He, Shuting and Bai, Song and Torr, Philip and Cao, Leilei and Zhang, Jinrong and Miao, Deshui and others},
  journal={arXiv preprint arXiv:2604.26031},
  year={2026}
}
\end{document}